%% file: main.tex
\documentclass[10pt,twocolumn,letterpaper]{article}

\usepackage[pagenumbers]{cvpr} %

\input{preamble}

\definecolor{cvprblue}{rgb}{0.21,0.49,0.74}
\usepackage[pagebackref,breaklinks,colorlinks,allcolors=cvprblue]{hyperref}

\def\confName{CVPR}
\def\confYear{2027}

\title{\method: Error-Resistant Autonomous Object Navigation with Predicted Depth}

\author{
Shunyang Huang$^{1*}$\quad Shuo Wang$^{2*}$\quad Wenyuan Xie$^{1}$\quad Mingyu Lin$^{1}$\quad Yue Ding$^{1,\dagger}$\quad Hongtao Lu$^{1,\dagger}$\\
$^{1}$Shanghai Jiao Tong University \qquad $^{2}$Tsinghua University
}

\begin{document}
\maketitle
{\renewcommand{\thefootnote}{}\footnotetext{$^*$Equal contribution. \quad $^\dagger$Corresponding authors.}}
\input{fig/teaser}
\input{sec/0_abstract}
\input{sec/1_intro}

\input{sec/2b_motivation}

\input{sec/3_method}

\input{sec/4_experiments}

\input{sec/2_related}
\input{sec/5_conclusion}
{
    \small
    \bibliographystyle{ieeenat_fullname}
    \bibliography{main}
}

\clearpage
\appendix
\crefalias{section}{appendix}
\crefalias{subsection}{appendix}
\twocolumn[\centering\Large\textbf{\thetitle}\\\vspace{0.5em}Appendix\\\vspace{1.0em}]
\input{sec/X_suppl}

\end{document}

%% file: preamble.tex
\usepackage{iftex}
\ifXeTeX\usepackage[OT1]{fontenc}\fi

\usepackage{cuted} %

\usepackage{currfile} %

\usepackage{caption} %

\usepackage{booktabs}   %
\usepackage{multirow}
\usepackage{xspace}
\usepackage{amssymb}
\usepackage{pifont}
\newcommand{\cmark}{\ding{51}}
\newcommand{\xmark}{\ding{55}}
\newcommand{\method}{ERA\xspace} %
\makeatletter
\newcommand{\res}[3]{\@ifundefined{res@#1@#2@#3}{\textbf{\color{red}??}}{\csname res@#1@#2@#3\endcsname}}
\newcommand{\abl}[2]{\@ifundefined{abl@#1@#2}{\textbf{\color{red}??}}{\csname abl@#1@#2\endcsname}}
\makeatother
\input{tables/numbers}

%% file: fig/teaser.tex
\begin{strip}
\centering
\includegraphics[width=\linewidth]{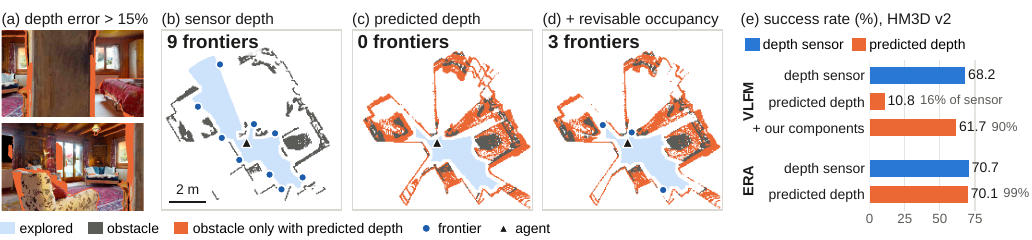}
\captionsetup{hypcap=false}
\captionof{figure}{\textbf{Effect of replacing the depth sensor with predicted depth.}
(a) Two views from VLFM's initial turn, highlighting pixels whose predicted depth differs from sensor depth by more than $15\%$.
(b--d) Maps at identical camera poses and their frontiers, used as exploration goals. Sensor depth leaves nine frontiers; accumulated predicted obstacles leave none; our revisable occupancy retains three using the same predictions.
(e) Success rates on HM3D v2. Our two components raise VLFM's success with predicted depth to $61.7\%$, while \method reaches $70.1\%$.}
\captionsetup{hypcap=true}
\label{fig:\currfilebase}%
\end{strip}

%% file: sec/0_abstract.tex
\begin{abstract}
Depth and 3D foundation models can predict geometry from RGB images, allowing zero-shot object navigation without a depth sensor.
However, when predicted depth is mapped as if it came from a sensor, view-dependent errors accumulate in the map, remove exploration goals, and end the search before the agent can observe and correct them.
We propose \method, a training-free object navigation system that replaces the depth sensor with depth predicted from RGB images and poses.
\method follows a paradigm we call \emph{Move with Error}: movement provides new views that correct map errors, and the corrected map enables further movement.
It realizes this paradigm with revisable occupancy and with exploration from reachable viewpoints, which continues the search when no frontier remains, and verifies the target across views before stopping.
On three benchmarks, \method achieves state-of-the-art performance among training-free methods without depth sensors.
Its map and exploration components also raise VLFM's success from $10.8\%$ to $61.7\%$ under the same predicted depth.
We further deploy \method on a real robot.
\end{abstract}

%% file: sec/1_intro.tex
\section{Introduction}
\label{sec:intro}

Zero-shot object navigation combines the semantic knowledge of foundation models with geometric maps to find objects in unseen environments~\cite{yokoyama2023vlfm,yin2024sgnav,yin2025unigoal,zhang2025apexnav}. These maps guide where to move, what to explore, and when to stop. Most map-based systems rely on depth sensors for geometry, but recent depth and 3D foundation models can predict geometry directly from RGB observations~\cite{lin2025da3,wang2025vggt}, enabling navigation without dedicated depth sensing. Existing RGB-only methods either reason in image space~\cite{cai2023pixnav,jin2025panonav,xie2026naviagent} or reconstruct geometry before planning~\cite{yu2025ranger,xie2026mvpnav}. This raises a central question: \emph{how should an agent navigate when the geometry on which it acts is inevitably imperfect?}

Replacing sensor depth with predicted depth can cause premature termination.
On HM3D v2, this substitution reduces VLFM's~\cite{yokoyama2023vlfm} success rate from $68.2\%$ to $10.8\%$, despite a mean relative depth error of about $10\%$ (\cref{fig:teaser}a); L3MVN~\cite{yu2023l3mvn}, which uses a different map and planner, also degrades substantially. Our analysis in \cref{sec:motivation} reveals how this failure arises: inconsistent predictions of the same surface accumulate into thicker obstacles, closing passages and eliminating frontiers---the boundaries between observed free space and unknown space (\cref{fig:teaser}b--d). Consequently, VLFM stops when no frontier remains, while L3MVN often stops when no exploration goal has an executable path. Once exploration stops, the agent can no longer acquire observations that might correct the map, leaving the errors that caused termination uncorrected.

This failure exposes a circular dependency: correcting the map requires new viewpoints, yet reaching those viewpoints is itself governed by the map.
We break this dependency with the \emph{Move with Error} paradigm: an agent should not treat the map only as a basis for navigation, but also use movement to acquire new observations that can test and revise its geometric understanding.
This paradigm requires two coupled capabilities. First, observations acquired through movement must be able to correct earlier geometric estimates when new viewpoints reveal contradictory evidence (\emph{Move2Error}). Second, the agent must continue seeking reachable viewpoints even when accumulated map errors eliminate its conventional exploration goals (\emph{Error2Move}).
Neither capability is sufficient on its own: map revision cannot help if the agent can no longer reach new viewpoints, while continued movement cannot correct the map if subsequent observations cannot revise earlier estimates.
Together, they establish a mutually reinforcing feedback loop in which movement enables map correction, and map correction enables further movement.

We instantiate Move with Error in \method, a training-free zero-shot object navigation system that uses RGB observations and pose estimates.
\emph{Revisable occupancy} lets later free-space observations remove erroneous obstacles while retaining geometry that no new evidence contradicts (Move2Error).
\emph{Exploration from reachable viewpoints} selects goals throughout reachable known-free space instead of only at frontiers, so that the agent keeps acquiring new views when no frontier remains (Error2Move).
The two mechanisms close the loop: exploration provides the observations needed for correction, and revisable occupancy turns them into map updates that reopen routes for further exploration.
\method further uses \emph{target verification}, which updates candidate geometry during approach and requires agreement across views before stopping.

By improving its resistance to errors during autonomous navigation, \method achieves state-of-the-art performance among training-free methods without depth sensors on three benchmarks.
The two mechanisms also transfer to existing systems: added to VLFM, they raise its success under the same predicted depth from $10.8\%$ to $61.7\%$ (\cref{fig:teaser}e) without changing its perception or controller; experiments with L3MVN and a second depth model further support their generality.
We also deploy \method on a real robot using a single RGB camera and wheel odometry to validate its effectiveness and robustness.

Our contributions are threefold:
\begin{itemize}[nosep,leftmargin=*]
  \item We analyze how predicted-depth errors propagate through mapping and exploration, revealing a failure in which geometric errors eliminate exploration goals and prevent the acquisition of further observations that could correct them.
  \item We introduce the \emph{Move with Error} paradigm and present \method as its instantiation, coupling revisable occupancy (Move2Error) with exploration from reachable viewpoints (Error2Move), complemented by target verification for the final approach and stopping decision.
  \item We demonstrate state-of-the-art performance among training-free methods without depth sensors on three benchmarks, and show that the two core mechanisms transfer to existing navigation systems and another depth predictor.
\end{itemize}

%% file: sec/2b_motivation.tex
\section{Problem Setting and Motivation}
\label{sec:motivation}
\label{sec:failure}

\subsection{Problem Setting}
\label{sec:setting}
Given a goal category $g$, the agent must find an instance in an unseen indoor environment using a single RGB camera.
At each step $t$, it receives an image $I_t$, camera intrinsics $K$, camera pose $T_t\in SE(3)$, planar position $x_t$, and collision feedback, then chooses to move forward, turn left or right, or call \textsc{Stop}.
Success requires stopping within the task's distance threshold of a goal instance.
No depth sensor is available, and all pretrained models remain frozen without navigation-specific training.

\subsection{Predicted Depth in Map-Based Systems}
\label{sec:pilot}

\input{tables/motivation}

We first examine how complete navigation systems behave when their depth sensor is replaced by predicted depth.
VLFM~\cite{yokoyama2023vlfm} explores frontiers in a 2D obstacle map, while L3MVN~\cite{yu2023l3mvn} ranks frontiers in a semantic map with a language model.
In each system, every depth-consuming component receives metric depth predicted by Depth Anything~3~\cite{lin2025da3} from RGB observations and camera poses; all other components are unchanged.

\paragraph{Observation.}
Both systems suffer a large drop in success rate (SR), as \cref{tab:motivation} shows.
Most failures involve premature termination of exploration: VLFM runs out of frontiers, while L3MVN often finds no path to an exploration goal.
The common failure is that the current map no longer provides an executable exploration goal.

\input{fig/maprules}

\paragraph{Diagnosis.}
To isolate the role of map updates, we compare rules along $120$ fixed trajectories.
All rules receive identical predicted-depth sequences, and we repeat the comparison with sensor depth at the same camera poses.
VLFM, SemExp~\cite{chaplot2020semexp}, and SG-Nav~\cite{yin2024sgnav} retain obstacle observations.
ApexNav~\cite{zhang2025apexnav} weighs obstacle and free-space evidence, while L3MVN clears cells observed as free in the current frame.
We also test a rebuilding rule that approximates the recent-frame map of MVP-Nav~\cite{xie2026mvpnav}, together with our revisable occupancy (\cref{sec:map}).

The comparisons show how view-dependent errors persist under different update rules.
Retaining every obstacle merges inconsistent surface predictions into thicker obstacles that can close passages, as illustrated by VLFM in \cref{fig:teaser}.
As \cref{fig:maprules}b and c show, ApexNav removes erroneous obstacles slowly because obstacle evidence outweighs free-space evidence, while L3MVN's current-view clearing leaves many earlier errors in place.
Rebuilding from recent frames discards both erroneous and real obstacles as old observations are removed, while errors still accumulate within each window.

These errors restrict exploration even when obstacles can be revised.
We measure whether a frontier can be reached from the agent through mapped free space.
As \cref{fig:maprules}a shows, every existing rule tested offers no reachable frontier in more than half of the predicted-depth frames, whereas this is uncommon with sensor depth.
The map can therefore lose exploration goals before the environment is fully explored, explaining the early termination observed in the complete systems.
Further analysis is provided in \cref{sec:analysis}.

\paragraph{Discussion.}
\label{sec:discussion}
The evidence points to two requirements, which together form \emph{Move with Error}.
First, observations acquired during movement must be able to revise earlier occupancy estimates (\emph{Move2Error}).
As \cref{fig:maprules}b and c show, revisable occupancy lets later free-space evidence remove erroneous obstacles while retaining most real ones.

Second, the agent must seek further observations when geometric errors obstruct exploration (\emph{Error2Move}).
Even the revisable map has no reachable frontier in almost half of the frames, as \cref{fig:maprules}a shows.
Ending the search at this point would prevent further correction.
Selecting goals throughout reachable free space, including previously mapped regions, preserves opportunities to obtain new views and revise the map.
The following method couples these two capabilities while leaving the depth predictor unchanged.

%% file: tables/motivation.tex
\begin{table}[t]
\centering
\begin{tabular}{@{}lcccc@{}}
\toprule
& \multicolumn{2}{c}{VLFM~\cite{yokoyama2023vlfm}} & \multicolumn{2}{c}{L3MVN~\cite{yu2023l3mvn}} \\
\cmidrule(lr){2-3}\cmidrule(lr){4-5}
Depth & SR & SPL & SR & SPL \\
\midrule
sensor                           & 68.2 & 34.5 & 63.2 & 27.5 \\
predicted                        & 10.8 & 7.5  & 29.2 & 13.7 \\
\bottomrule
\end{tabular}
\caption{\textbf{Replacing the depth sensor by predicted depth} in two zero-shot navigation systems on HM3D v2. SR: success rate; SPL: success weighted by path length (both in \%).}
\label{tab:motivation}
\end{table}

%% file: fig/maprules.tex
\begin{figure}[t]
  \centering
  \includegraphics[width=\linewidth]{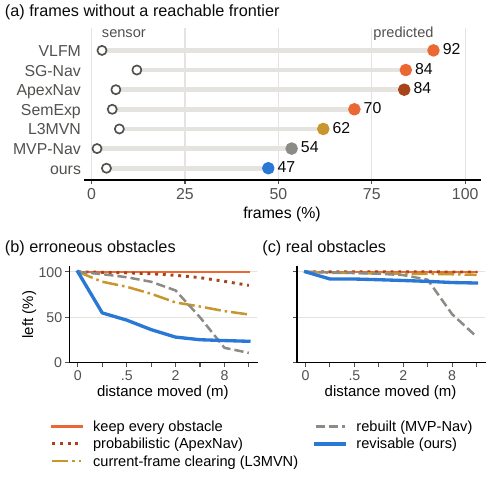}
  \caption{\textbf{Map update rules under predicted depth}, along the same $120$ trajectories on HM3D v2.
  (a) Frames in which the map offers no reachable frontier, with sensor and with predicted depth.
  (b, c) Erroneous and real obstacles that remain in a map built from predicted depth as the agent moves on; VLFM, SemExp and SG-Nav keep every obstacle.}
  \label{fig:maprules}
\end{figure}

%% file: sec/3_method.tex
\section{Method}
\label{sec:method}

\input{fig/overview}

\method instantiates the \emph{Move with Error} paradigm as a closed-loop navigation process, as illustrated in \cref{fig:overview}.
Each new view updates an occupancy map, a semantic value map~\cite{yokoyama2023vlfm}, and object candidates. This loop does not depend on the availability of frontiers: when predicted-depth errors eliminate conventional exploration goals, \method continues selecting viewpoints throughout reachable known-free space, where new views may reveal free space behind erroneous obstacles. \emph{Revisable occupancy} (Move2Error, \cref{sec:map}) uses this evidence to correct earlier obstacle estimates, while \emph{exploration from reachable viewpoints} (Error2Move, \cref{sec:explore}) ensures that the agent can continue acquiring such corrective observations. Once a target candidate is confirmed, the agent switches to approach and verification, updating its surface estimate and checking it across views before stopping (\cref{sec:target}).

The geometric input is metric depth $\hat D_t$ and confidence $C_t$ from frozen Depth Anything~3~\cite{lin2025da3}, given the current image, recent keyframes, intrinsics, and poses.
The poses constrain prediction across views; the model predicts metric scale.
Frozen vision--language and segmentation models provide goal-conditioned image--text similarity and masks for the goal and visually similar categories.
The depth predictor remains unchanged throughout navigation.

\subsection{Revisable Occupancy}
\label{sec:map}

\input{fig/clearing}

Move2Error requires new observations to overturn an earlier obstacle estimate when they reveal free space at the same location.
Confidence filtering alone cannot provide this revision: it can miss smooth depth biases, and accepted errors still remain in the map.
Building on classic occupancy mapping~\cite{elfes1989occupancy,hornung2013octomap}, we integrate obstacle and free-space observations as bounded evidence, retaining earlier geometry where no new evidence is available.

\paragraph{Bounded evidence and free-space clearing.}
Each grid cell $m$ stores signed evidence $e_m\in[-E,E]$, initially zero, and an observed flag to distinguish unknown cells from free space.
Back-projecting $\hat D_t$ marks cells containing points within the agent's body-height range as obstacle observations.
Free-space rays in each group of image columns end before predicted obstacles, or at observed floor when none is present.
Traversed cells receive free-space observations.
We update evidence by
\begin{equation}
 e_m\leftarrow
 \begin{cases}
  \min\!\big(E,\;\max(0,e_m)+\delta\big) & \text{obstacle observation},\\[2pt]
  \max\!\big(-E,\;e_m-\delta\big) & \text{free observation},\\[2pt]
  e_m & \text{otherwise},
 \end{cases}
 \label{eq:evidence}
\end{equation}
where $E>0$ bounds the evidence and $\delta>0$ is the update magnitude.
A cell is predicted occupied when $e_m>0$; obstacle observations take precedence over free-space observations within a frame.

The asymmetric obstacle update gives a new obstacle positive evidence immediately, even if the cell was previously free.
The upper bound ensures that repeated predictions cannot accumulate unlimited obstacle evidence.
With $\delta=E/2$, one subsequent free observation clears a cell supported by one obstacle observation, and two consecutive free observations clear saturated evidence.
This \emph{free-space clearing} revises occupancy when later views contradict an earlier obstacle, as illustrated in \cref{fig:clearing}.
\Cref{fig:maprules}b and c evaluate its effect on erroneous and real obstacles.

\paragraph{Motion feedback and traversable space.}
Executed motion provides evidence independent of predicted depth.
After every map update, cells previously swept by the agent's footprint are reset to $e_m=-E$, preserving physically traversed routes against later contradictory predictions.
When a forward action fails, cells around its intended destination enter a contact-obstacle set $\mathcal K_t$, following Stubborn~\cite{luo2022stubborn}.
These cells cannot be cleared by visual evidence, only by the agent's center passing through them.
Let $\mathcal O_t$ denote observed cells and $\mathcal B_t=\{m:e_m>0\}\cup\mathcal K_t$ denote blocked cells.
The traversable space used for planning is
\begin{equation}
 \mathcal F_t=\{m\in\mathcal O_t:\ \operatorname{dist}(m,\mathcal B_t)>r\},
 \label{eq:traversable}
\end{equation}
where $r$ is the agent radius and $\operatorname{dist}$ measures distance to the nearest blocked cell.
Thus, planning uses observed free space with clearance for the agent, and revising a predicted obstacle can reopen a route through it.

\subsection{Exploration from Reachable Viewpoints}
\label{sec:explore}

Error2Move addresses the remaining obstacle to map revision: obtaining another view when the current map offers no frontier.
Frontier-based exploration~\cite{yamauchi1997frontier,yokoyama2023vlfm} selects goals at boundaries between observed free space and unknown space, which can disappear prematurely under predicted depth even with revisable occupancy.
\method instead draws its exploration goals from reachable known-free space throughout the search.
This permits movement within mapped regions, where a different view may reveal free space behind an erroneous obstacle.

\paragraph{Viewpoint selection.}
Let $\rho_t(v)$ be the shortest route length from the current position $x_t$ to cell $v$ through the traversable space $\mathcal F_t$, with $\rho_t(v)=\infty$ for unreachable cells.
Instead of requiring $v$ to border unknown space, \method selects the next viewpoint from the candidate viewpoints $\mathcal V_t$, the cells of $\mathcal F_t$ whose route length lies between $\rho_{\min}$ and $\rho_{\max}$:
\begin{equation}
 \begin{gathered}
 \mathcal V_t=\{v\in\mathcal F_t:\ \rho_{\min}\le\rho_t(v)\le\rho_{\max}\},\\
 v_t^*=\operatorname*{arg\,max}_{v\in\mathcal V_t}\;G(v)-C(v).
 \end{gathered}
 \label{eq:candidates}
\end{equation}
The gain $G(v)$ combines the unknown space visible from $v$, the goal-conditioned image--text similarity stored in a value map as in VLFM~\cite{yokoyama2023vlfm}, and how much of the region containing $v$ is still uncovered; the cost $C(v)$ penalizes long routes, small obstacle clearance, and repeated visits.
$\mathcal V_t$ is sampled on a coarse grid, and both terms are defined in \cref{sec:suppl_explore}.

Visible unknown space only ranks the candidates and does not decide which viewpoints are eligible.
When no candidate sees unknown space, \method still selects a viewpoint in mapped space, and its next observation can clear an erroneous obstacle through \cref{eq:evidence} and reopen a route.
This couples Error2Move with Move2Error without requiring the agent to identify in advance which obstacle is erroneous.

\paragraph{Execution and termination.}
\label{sec:executor}
To avoid oscillation as the map changes, \method retains a selected goal until it is reached, becomes unreachable, progress stalls, or its action budget expires.
A distance-field planner on the traversable space $\mathcal F_t$ converts the goal into short sequences of discrete actions.
If $\mathcal V_t$ is empty, the agent turns in place to acquire further observations; exploration is declared exhausted only if no candidate becomes available over three full rotations.
When transferring the components to existing systems, we replace their map update but invoke reachable-viewpoint selection only when their original exploration cannot supply an executable goal.
The transfer procedure is detailed in \cref{sec:suppl_transfer}.

\subsection{Target Verification}
\label{sec:target}

Target verification complements exploration by updating candidate geometry during approach and checking it before \textsc{Stop}.
Each candidate $c=(b_c,P_c,\mathcal H_c)$ stores a semantic score $b_c$, its latest reliable surface $P_c$, and an observation history $\mathcal H_c$ with camera poses.
We split goal masks into depth-connected surfaces using pixels with sufficient confidence $C_t$ and consistency under reprojection from the previous view.
These reliability checks apply to target surfaces, while occupancy uses the revision rule in \cref{eq:evidence}.

\paragraph{Candidate update and approach.}
A surface is associated with an existing candidate by 3D overlap.
If this fails, image overlap can establish a match only when the surfaces also lie within an uncertainty-dependent 3D tolerance.
In every frame, the semantic score of each candidate is updated as
\begin{equation}
 b_c\leftarrow\operatorname{clip}_{[-b_{\max},\,b_{\max}]}\big(b_c+w(\ell_c^+-\ell_c^-)+a_c-m_c\big),
 \label{eq:score_update}
\end{equation}
where $\ell_c^+$ and $\ell_c^-$ are bounded strengths of the strongest goal and similar-category detections on the candidate, $w$ weights them, $a_c$ is a small term that compares the image--text similarity of the candidate with the goal and with similar categories, $m_c$ penalizes a clear view of the candidate without a goal detection, and $b_{\max}$ bounds the score.
When a surface $S_{c,t}$ is matched to $c$ with enough reliable points, it replaces $P_c$ and is added to $\mathcal H_c$ together with the camera pose.
Semantic evidence accumulates, but the surface estimate is replaced, not fused with earlier surfaces, which remain in the history for verification.
This follows our analysis in \cref{sec:suppl_error}: nearby views have correlated depth errors, while approaching reduces their absolute magnitude.
For a candidate with sufficient semantic support and plausible geometry, \method plans to a reachable endpoint near its surface and keeps the target visible during the final approach.

\paragraph{Verification before stopping.}
Let $d_t$ be the planar distance to a locally supported part of the target surface that is depth-consistent with the current view, and $\tau$ the stopping threshold.
\method calls \textsc{Stop} only when $d_t<\tau$ and four checks pass: the candidate is confirmed, detected in the current frame, and favored over similar categories; the agent faces the target; the surface has sufficient support with low conflict and uncertainty; and an earlier observation from a different camera position agrees with the current one.
If geometry or multi-view support is insufficient, \method moves to another reachable position and turns back toward the object to obtain an additional observation.
It returns to exploration if the verification budget expires.
Association, approach, and stopping checks are detailed in \cref{sec:suppl_target}.

%% file: fig/overview.tex
\begin{figure*}[t]
  \centering
  \includegraphics[width=\linewidth]{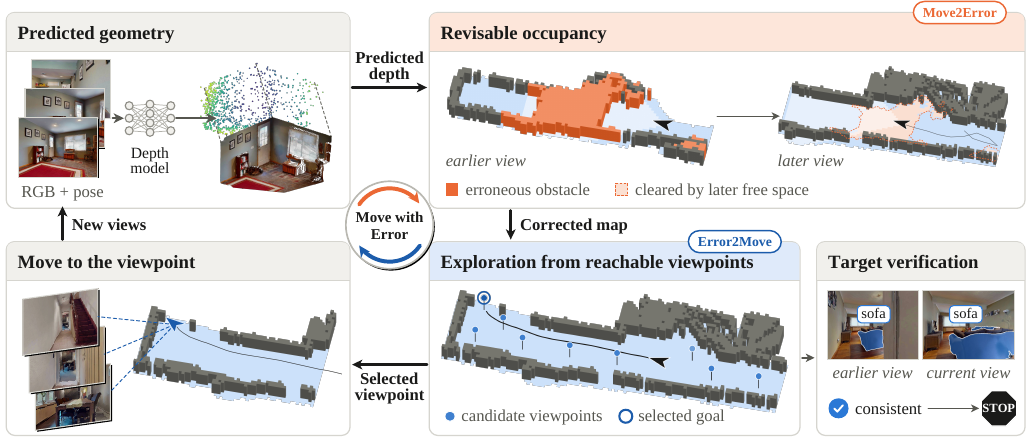}
  \caption{\textbf{Overview of \method.}
    Depth predicted by a frozen model from posed RGB images can add erroneous obstacles to the map.
    \emph{Revisable occupancy} (Move2Error) removes them when later views observe free space there.
    \emph{Exploration from reachable viewpoints} (Error2Move) selects the next viewpoint in reachable free space, even when no frontier remains, and moving there yields new views.
    \emph{Target verification} checks a target candidate in views from different positions before stopping.
    The panels are illustrations from real maps and views.}
  \label{fig:overview}
\end{figure*}

%% file: fig/clearing.tex
\begin{figure}[t]
  \centering
  \includegraphics[width=\linewidth]{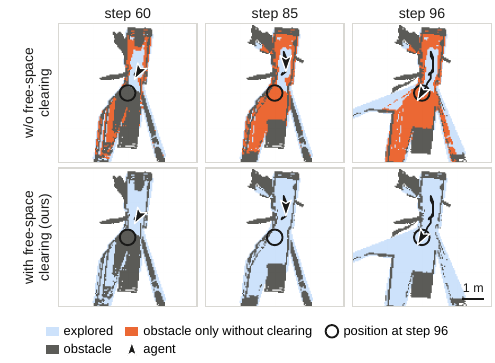}
  \caption{\textbf{Free-space clearing.}
    The same observations are fused into a map without (top) and with (bottom) free-space clearing.
    Predicted depth places an obstacle across the passage (step 60).
    Later views clear it in our map (step 85), and the agent walks through (step 96); without clearing, the obstacle stays and the passage remains closed.}
  \label{fig:clearing}
\end{figure}

%% file: sec/4_experiments.tex
\section{Experiments}
\label{sec:experiments}

\input{tables/main}

\subsection{Experimental Setup}
\label{sec:setup}

\paragraph{Benchmarks and metrics.}
We evaluate \method on three widely used ObjectNav benchmarks in the Habitat simulator~\cite{savva2019habitat}: HM3D v1 and HM3D v2~\cite{ramakrishnan2021hm3d,yadav2022hm3dsem} and MP3D~\cite{chang2017mp3d}.
HM3D v1 provides 2{,}000 validation episodes in 20 scenes and HM3D v2 1{,}000 episodes in 36 scenes, both with six goal categories, while MP3D provides 2{,}195 validation episodes in 11 scenes with 21 categories.
Following~\cite{batra2020objectnavrevisited}, we report the Success Rate (SR), which is the fraction of episodes in which the agent calls \textsc{Stop} within $1$\,m of the goal object, and the Success weighted by Path Length (SPL), which additionally measures how efficient the successful paths are.

\paragraph{Implementation details.}
The agent observes $640\times480$ RGB images with a horizontal field of view of $79^\circ$ and moves with discrete actions ($0.25$\,m forward, $30^\circ$ turns) for at most $500$ steps per episode.
We use Depth Anything~3~\cite{lin2025da3}, in its nested giant-large variant (DA3NESTED-GIANT-LARGE), for depth prediction, and C-RADIOv4~\cite{ranzinger2026cradiov4} with its SigLIP2 adaptor~\cite{tschannen2025siglip2} and the SAM\,3 head~\cite{carion2025sam3} for perception.
All models are used off the shelf without fine-tuning, and all experiments run on two NVIDIA A100 GPUs.
The poses are those provided by the simulator; the sensitivity of \method to odometry drift is examined in \cref{sec:suppl_drift}.
Further details are given in \cref{sec:suppl_explore}.

\subsection{Comparison with State-of-the-Art Methods}
\label{sec:main}

\input{tables/generalization}

\Cref{tab:main} compares \method with methods trained for navigation, training-free methods that rely on a depth sensor, and training-free methods that do not.
Among the latter, \method achieves the best SR and SPL on all three benchmarks.
Among the other methods in this group, MVP-Nav~\cite{xie2026mvpnav} is the strongest; \method exceeds its SR by $10.7$ points on HM3D v2 and by smaller margins on HM3D v1 and MP3D.
It is also competitive with training-free methods that use a depth sensor: on the two HM3D benchmarks only VLFM~\cite{yokoyama2023vlfm} reaches a higher SPL, and on MP3D none does.

\subsection{Generalization across Systems and Depth Sources}
\label{sec:generalization}
\label{sec:depthsource}

\Cref{tab:generalization} examines whether our findings hold beyond our own system and beyond the depth model used so far.
We add revisable occupancy and exploration from reachable viewpoints to the two systems of \cref{sec:pilot} without changing their depth input, perception or controller.
In VLFM, revisable occupancy raises the SR under predicted depth to $50.7\%$, and continuing toward a reachable viewpoint where VLFM would otherwise stop raises it to $61.7\%$, $90\%$ of what VLFM achieves with a depth sensor.
L3MVN~\cite{yu2023l3mvn} builds a different map and ends its search when its planner finds no path to an exploration goal; revisable occupancy alone raises its SR from $29.2\%$ to $54.8\%$, and both components to $59.1\%$.
Neither change makes the depth more accurate; both only change how the system uses it.
When DA3 is replaced by VGGT~\cite{wang2025vggt}, whose up-to-scale predictions are converted to metric depth with the known camera height and which is less accurate than DA3 in this setting, VLFM and L3MVN again collapse, to $8.3\%$ and $26.5\%$, our two components restore them to $59.8\%$ and $60.6\%$, and \method reaches $69.5\%$, so neither the failure nor the effect of our components is specific to one depth model.
Details and example trajectories are given in \cref{sec:suppl_transfer,sec:suppl_vggt}.

\subsection{Ablation Study}
\label{sec:ablation}

\input{tables/ablation}

\Cref{tab:ablation} shows the contribution of each component of \method on HM3D v2.
When free-space rays no longer clear occupied cells, the SR drops from \res{v2}{1.0}{SR}\% to $13.9\%$: erroneous obstacles stay in the map and gradually confine the agent, which both hides the goal and blocks the way to it.
Restricting the exploration goals to frontiers reduces the SR to $32.6\%$, although the map remains revisable, since most episodes again end without any goal candidate; with sensor depth, the same restriction costs less than five points, so reachable viewpoints become necessary when the depth is predicted.
Without target verification, candidate positions are no longer updated across views and the agent stops without the consistency and multi-view checks; the SR then decreases by $2.5$ points.
A breakdown is given in \cref{sec:suppl_target_ablation}.

\subsection{Real-World Deployment}
\label{sec:realworld}

\input{fig/realworld}

We deploy \method on an AgiBot Genie G1 wheeled humanoid robot, where it receives only the images of the RGB camera in the head of the robot and the poses of its wheel odometry; none of its depth sensors is used.
In the corridors of a university building, the robot turns in place to observe its surroundings, navigates toward the goal object once it is detected, and stops on its own, as shown in \cref{fig:realworld}.
We run it $12$ times with three object categories, chair, trash can and plant; in seven runs, the target is not in view at the start.
As \cref{tab:real} shows, the robot stops within $1$\,m of the target in $11$ of them; all runs are listed in \cref{sec:suppl_real}.

To test the two components on the robot, we start from a base that, like VLFM, never removes obstacles from the map and selects its exploration goals among frontiers only, and add revisable occupancy and then reachable viewpoints.
All three versions start from the same three positions for each of two placements of a trash can.
As the lower part of \cref{tab:real} shows, the base ends every run because no reachable frontier is left, and revisable occupancy alone succeeds once; only with reachable viewpoints does \method reach the target in all six runs, and the bottom row of \cref{fig:realworld} compares the base with \method from one start.

\begin{table}[t]
\centering
\setlength{\tabcolsep}{4.5pt}
\begin{tabular}{@{}llcc@{}}
\toprule
Runs & System & Success & SR \\
\midrule
Three categories & \method & 11 / 12 & 91.7 \\
\midrule
Same starts & Base & 0 / 6 & 0 \\
 & + revisable occupancy & 1 / 6 & 16.7 \\
 & + reachable viewpoints & 6 / 6 & 100 \\
\bottomrule
\end{tabular}
\caption{\textbf{Real-robot runs.} Success counts the runs that stop within $1$\,m of the target. The lower part compares runs from the same starts. Base uses the map update and frontier goals of VLFM; each row adds one component to the row above, and the last row is \method.}
\label{tab:real}
\end{table}

%% file: tables/main.tex
\begin{table*}[t]
\centering
\begin{tabular}{@{}lcc cc cc cc@{}}
\toprule
& & & \multicolumn{2}{c}{HM3D v1} & \multicolumn{2}{c}{HM3D v2} & \multicolumn{2}{c}{MP3D} \\
\cmidrule(lr){4-5}\cmidrule(lr){6-7}\cmidrule(lr){8-9}
Method & Depth sensor & Training-free & SR & SPL & SR & SPL & SR & SPL \\
\midrule
SemExp~\cite{chaplot2020semexp}      & \cmark & \xmark & -- & -- & -- & -- & 36.0 & 14.4 \\
ZSON~\cite{majumdar2022zson}         & \xmark & \xmark & 25.5 & 12.6 & -- & -- & 15.3 & 4.8 \\
PixNav~\cite{cai2023pixnav}          & \xmark & \xmark & 37.9 & 20.5 & -- & -- & -- & -- \\
\midrule
ESC~\cite{zhou2023esc}               & \cmark & \cmark & 39.2 & 22.3 & -- & -- & 28.7 & 14.2 \\
L3MVN~\cite{yu2023l3mvn}             & \cmark & \cmark & 50.4 & 23.1 & 63.2$^\dagger$ & 27.5$^\dagger$ & 34.9 & 14.5 \\
VoroNav~\cite{wu2024voronav}         & \cmark & \cmark & 42.0 & 26.0 & -- & -- & -- & -- \\
OpenFMNav~\cite{kuang2024openfmnav}  & \cmark & \cmark & 54.9 & 24.4 & -- & -- & -- & -- \\
VLFM~\cite{yokoyama2023vlfm}        & \cmark & \cmark & 52.5 & 30.4 & 68.2$^\dagger$ & 34.5$^\dagger$ & 36.4 & 17.5 \\
SG-Nav~\cite{yin2024sgnav}           & \cmark & \cmark & 54.0 & 24.9 & 49.6 & 25.5 & 40.2 & 16.0 \\
InstructNav~\cite{long2024instructnav} & \cmark & \cmark & -- & -- & 58.0 & 20.9 & -- & -- \\
UniGoal~\cite{yin2025unigoal}        & \cmark & \cmark & 54.5 & 25.1 & -- & -- & 41.0 & 16.4 \\
ImagineNav~\cite{zhao2024imaginenav} & \cmark & \cmark & 53.0 & 23.8 & -- & -- & -- & -- \\
\midrule
PanoNav~\cite{jin2025panonav}        & \xmark & \cmark & 43.5 & 23.7 & -- & -- & -- & -- \\
RANGER~\cite{yu2025ranger}           & \xmark & \cmark & -- & -- & 42.7 & 17.8 & -- & -- \\
MVP-Nav$^\dagger$~\cite{xie2026mvpnav} & \xmark & \cmark & 49.6 & 24.3 & 59.4 & 25.6 & 38.6 & 17.6 \\
\textbf{\method (ours)}              & \xmark & \cmark & \textbf{\res{v1}{1.0}{SR}} & \textbf{\res{v1}{1.0}{SPL}} & \textbf{\res{v2}{1.0}{SR}} & \textbf{\res{v2}{1.0}{SPL}} & \textbf{\res{mp3d}{1.0}{SR}} & \textbf{\res{mp3d}{1.0}{SPL}} \\
\bottomrule
\end{tabular}
\caption{\textbf{Object navigation on the validation splits} of HM3D v1, HM3D v2 and MP3D (SR / SPL in \%). $^\dagger$Evaluated under our protocol. Bold: best among training-free methods without a depth sensor.}
\label{tab:main}
\end{table*}

%% file: tables/generalization.tex
\begin{table}[t]
\centering
\setlength{\tabcolsep}{2.5pt}
\begin{tabular}{@{}lcccccc@{}}
\toprule
& \multicolumn{2}{c}{Sensor} & \multicolumn{2}{c}{DA3} & \multicolumn{2}{c}{VGGT} \\
\cmidrule(lr){2-3}\cmidrule(lr){4-5}\cmidrule(lr){6-7}
System & SR & SPL & SR & SPL & SR & SPL \\
\midrule
VLFM~\cite{yokoyama2023vlfm} & 68.2 & 34.5 & 10.8 & 7.5  & 8.3  & 6.1 \\
\quad + revisable occupancy  & --   & --   & 50.7 & 25.8 & 52.4 & 28.5 \\
\quad + reachable viewpoints  & --   & --   & 61.7 & 30.2 & 59.8 & 31.5 \\
\midrule
L3MVN~\cite{yu2023l3mvn}     & 63.2 & 27.5 & 29.2 & 13.7 & 26.5 & 13.2 \\
\quad + revisable occupancy  & --   & --   & 54.8 & 24.3 & 56.3 & 25.3 \\
\quad + reachable viewpoints  & --   & --   & 59.1 & 25.0 & 60.6 & 26.2 \\
\midrule
\method                      & 70.7 & 33.7 & \res{v2}{1.0}{SR} & \res{v2}{1.0}{SPL} & 69.5 & 30.5 \\
\bottomrule
\end{tabular}
\caption{\textbf{Generalization} on HM3D v2 across systems and depth sources. The rows below each system are cumulative. Implementation details are given in \cref{sec:suppl_transfer}.}
\label{tab:generalization}
\label{tab:depth}
\end{table}

%% file: tables/ablation.tex
\begin{table}[t]
\centering
\begin{tabular}{@{}lcc@{}}
\toprule
Variant & SR & SPL \\
\midrule
\method (full)                      & \res{v2}{1.0}{SR} & \res{v2}{1.0}{SPL} \\
\quad w/o free-space clearing      & 13.9 & 6.4 \\
\quad frontier goals only           & 32.6 & 18.1 \\
\quad w/o target verification       & 67.6 & 29.9 \\
\midrule
\multicolumn{3}{@{}l}{\emph{With sensor depth}} \\
\method (full)                      & 70.7 & 33.7 \\
\quad frontier goals only           & 66.2 & 33.9 \\
\bottomrule
\end{tabular}
\caption{\textbf{Ablation study} on HM3D v2. Each variant changes one component of the full system.}
\label{tab:ablation}
\end{table}

%% file: fig/realworld.tex
\begin{figure}[t]
  \centering
  \includegraphics[width=\linewidth]{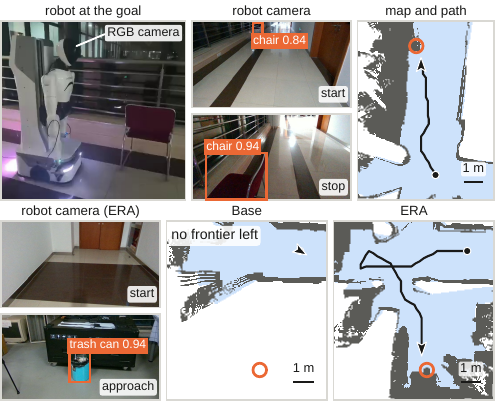}
  \caption{\textbf{Real-world deployment.}
    \method on a wheeled humanoid robot that uses only the RGB camera in its head.
    Top, from left to right: the robot at the goal, the view of its camera at the start and at the stop with the detection of the goal, and the map built from predicted depth with the path of the robot (black) and the goal (orange circle).
    Bottom: two runs from the same start, Base (\cref{tab:real}) and \method, and their maps at the end.}
  \label{fig:realworld}
\end{figure}

%% file: sec/2_related.tex
\section{Related Work}
\label{sec:related}

\paragraph{Zero-shot object navigation.}
Early object navigation agents learn to build and explore semantic maps~\cite{chaplot2020semexp,ramakrishnan2022poni} or to map observations directly to actions~\cite{majumdar2022zson,yadav2023ovrlv2}.
Training-free methods use foundation models to guide exploration through vision--language similarity~\cite{yokoyama2023vlfm}, language-model reasoning~\cite{zhou2023esc,yu2023l3mvn,wu2024voronav,kuang2024openfmnav,long2024instructnav}, scene graphs~\cite{yin2024sgnav,yin2025unigoal}, world models~\cite{nie2025wmnav}, or adaptive exploration strategies~\cite{zhang2025apexnav}.
Many of these advances share a geometric backbone: a map built from sensor depth supports path planning and frontier-based exploration~\cite{yamauchi1997frontier}.
We study how this geometric backbone should change when depth is predicted and its errors affect subsequent exploration.

\paragraph{Navigation without a depth sensor.}
One line of work avoids metric maps and reasons in image space: PixNav~\cite{cai2023pixnav} specifies navigation goals as pixels, while PanoNav~\cite{jin2025panonav} selects directions from panoramic scene descriptions. Navi-Agent~\cite{xie2026naviagent} follows language instructions with a topology of visual places connected by executed motions, without depth or global coordinates. ImagineNav~\cite{zhao2024imaginenav} similarly selects imagined views, although it still uses depth for low-level control.
These representations reduce dependence on explicit geometry, but do not directly encode metric free space and reachability over long horizons.
Another line uses visual geometry models such as DUSt3R, VGGT, and Depth Anything~3~\cite{wang2023dust3r,wang2025vggt,lin2025da3} to reconstruct geometry for planning. RANGER~\cite{yu2025ranger} fuses keyframe reconstructions into an occupancy map, where errors can accumulate across views; MVP-Nav~\cite{xie2026mvpnav} rebuilds its local map from recent frames, limiting this accumulation while discarding earlier geometry.
The effect of estimated depth on navigation has also been studied in classic pipelines~\cite{mishkin2019benchmarking} and ObjectNav~\cite{goetting2024vlmnav}; FrontierNet~\cite{sun2025frontiernet} learns exploration goals directly from images to avoid extracting them from unreliable depth.
We retain explicit geometry and accept that its predictions remain imperfect, focusing on how navigation can use subsequent observations to revise map errors.

\paragraph{Occupancy mapping and exploration.}
Classic occupancy mapping integrates obstacle returns and free-space rays with bounded evidence~\cite{elfes1989occupancy,hornung2013octomap}, allowing later observations to revise a cell.
Several map-based navigation systems retain obstacle updates throughout an episode~\cite{yokoyama2023vlfm,chaplot2020semexp,yin2024sgnav}; ApexNav~\cite{zhang2025apexnav} uses probabilistic updates that give obstacle evidence substantially more weight than free-space evidence. Under predicted depth, these choices can make view-dependent errors persist.
Our revisable occupancy builds on the classic formulation and incorporates action feedback as in Stubborn~\cite{luo2022stubborn}. Selecting viewpoints by the unknown space they reveal follows next-best-view planning~\cite{bircher2016nbv}; here, reachable mapped space also supplies new observations when no frontier remains.
As in recent methods~\cite{zhang2025apexnav,wu2026rememnav,pei2026sapnav,liang2026zonda}, we further verify targets before stopping, using observations across views of predicted geometry.

%% file: sec/5_conclusion.tex
\section{Conclusion}
\label{sec:conclusion}
In this paper, we studied zero-shot object navigation with depth predicted from RGB images instead of a depth sensor.
We showed that view-dependent errors of predicted depth accumulate in maps built as if from a depth sensor, remove exploration goals, and end the search before the agent can correct them.
To address this, we proposed the \emph{Move with Error} paradigm, in which movement corrects map errors and the corrected map enables further movement, and instantiated it in \method with revisable occupancy, exploration from reachable viewpoints, and multi-view target verification.
\method achieves state-of-the-art results among training-free methods without depth sensors on three benchmarks and runs on a real robot.
Added to two existing systems, its map and exploration components recover most of their success lost under predicted depth, with either of two depth models.
\method relies on the poses provided by the platform, and continuing the search from mapped space can lengthen the path to the goal; we leave pose estimation and more efficient viewpoint selection to future work.

%% file: sec/X_suppl.tex
\makeatletter\setlength{\@dblfptop}{0pt}\makeatother

\section{Analysis of Exploration under Predicted Depth}
\label{sec:analysis}

\subsection{Why the Map of VLFM Closes}
\label{sec:suppl_replay}
\input{fig/suppl_replay}
We first take VLFM, the system of \cref{fig:teaser}, as a representative case, and check that its failure with predicted depth comes from its map and not from a different trajectory; \cref{sec:suppl_rules} then compares the maps of other systems.
In $120$ HM3D v2 episodes, we follow the trajectory of VLFM with predicted depth and run its mapping and frontier extraction twice at the same poses, once with sensor depth and once with DA3 depth.
The two maps therefore differ only in the depth they receive.
As \cref{fig:suppl_replay}c shows, the predicted-depth map loses all of its frontiers in $114$ of the $120$ episodes, half of them already during the initial turn, and it completely encloses the agent's traversable region in $108$ episodes, with a median enclosed area of $6.5$\,m$^2$.
With sensor depth, neither happens in a single episode.
Most of the additional obstacle cells lie next to real obstacles: $68\%$ of them are within $0.25$\,m of a real obstacle and fewer than $1\%$ are more than $1$\,m away.
As \cref{fig:suppl_replay}d shows, most of them come from depth edges, where the prediction is smeared between the near and the far surface, and from small offsets of otherwise correct surfaces.
Few come from far regions beyond the range of the sensor.
Over the views of the initial turn, these cells thicken the real obstacles until the gaps between them close, as in \cref{fig:suppl_replay}b.

The predicted depth itself is fairly accurate.
We compare it with the sensor depth on the same $2{,}595$ frames.
The error of a frame is the mean relative error over the pixels whose sensor depth lies between $0.5$ and $5$\,m.
It is $9.9\%$ on average, $6.7\%$ in the median frame and $18.3\%$ at the 90th percentile; this is the error of about $10\%$ quoted in \cref{sec:intro}.

\subsection{Map Update Rules: Details}
\label{sec:suppl_rules}
\input{fig/suppl_maprules}
This section defines the quantities of \cref{fig:maprules}, which compares how the maps of existing systems treat the errors of predicted depth.
The episodes are the $120$ of \cref{sec:suppl_replay}, but the trajectories are those of \method with predicted depth, which keep moving through the scene instead of ending early, so that every map is observed over the whole scene.
Along each, we build the obstacle map of VLFM, the projection maps of SemExp~\cite{chaplot2020semexp} and SG-Nav~\cite{yin2024sgnav} and the probabilistic map of ApexNav~\cite{zhang2025apexnav} and the map of L3MVN~\cite{yu2023l3mvn} with their released code and parameters on a common fixed grid, and revisable occupancy, once from predicted and once from sensor depth.
The map of ApexNav updates the occupancy probability of a cell once per frame; with its parameters, an obstacle that has been observed once needs $28$ frames of free-space observations before it is removed, and we examine stronger settings below.
L3MVN keeps for every cell the largest value over all frames and then clears the cells that the current frame observes as free, except those next to an obstacle in the current frame.
The rebuilt map replaces the map, every $30$ frames, by the one built from the last $30$ frames, which approximates the rule of MVP-Nav~\cite{xie2026mvpnav} without its reconstruction network.
\paragraph{Reachable frontiers.}
\Cref{fig:maprules}a counts the frames in which frontier-based exploration would have no goal.
The observed space is the region that the depth of all frames so far has covered, up to the farthest valid point in every direction.
A frontier is a connected set of at least $10$ observed cells next to unobserved space that the agent can reach in the map, with obstacles enlarged by the agent radius; cells within $0.5$\,m of the agent are ignored, and the frames of the initial turn are left out.
\Cref{fig:suppl_maprules} shows the maps of all rules at one frame of an episode in which every rule still has a reachable frontier with sensor depth.
Among the frames in which the map built from sensor depth has a reachable frontier, the map built from predicted depth has none in $91\%$ of them for VLFM, $83\%$ for ApexNav, $82\%$ for SG-Nav, $69\%$ for SemExp, $59\%$ for L3MVN, $53\%$ for the rebuilt map and $45\%$ for revisable occupancy.

\paragraph{Tuning ApexNav.}
Since the slow removal in the map of ApexNav comes from its parameters, we also run it with free-space observations weighted more strongly, so that an obstacle observed many times is removed after $16$, $8$, $4$ or $2$ frames of free-space observations instead of $32$.
As \cref{tab:suppl_apex} shows, faster removal brings the reachable frontiers of this map close to those of revisable occupancy, but about half of its erroneous obstacles remain, against a quarter in revisable occupancy, and the strongest setting also removes more real obstacles.
Revisable occupancy differs in which cells receive free-space observations, along the rays to the nearest obstacle or to the floor, and in clearing the cells swept by the agent, which this map of ApexNav does not do and which accounts for part of the difference in enclosure.
Every setting still leaves the agent without a reachable frontier in more than a third of the frames, so exploration cannot rely on frontiers even with a better map.

\begin{table}[h]
\centering
\small
\setlength{\tabcolsep}{3.4pt}
\begin{tabular}{@{}lccccc@{}}
\toprule
Map & Frames & Error & Real & Encl. & No frontier \\
\midrule
ApexNav (released) & 32 & 86.6 & 99.8 & 28.6 & 82.6 \\
ApexNav & 16 & 68.2 & 99.1 & 13.9 & 75.3 \\
ApexNav & 8 & 53.5 & 96.9 & 6.0 & 60.9 \\
ApexNav & 4 & 46.0 & 92.3 & 4.9 & 47.4 \\
ApexNav & 2 & 46.2 & 84.5 & 4.7 & 38.1 \\
Revisable occ. & 2 & 24.3 & 88.8 & 0.0 & 45.1 \\
\bottomrule
\end{tabular}
\caption{\textbf{ApexNav with stronger free-space updates}, built from predicted depth along the same $120$ trajectories (\%). \emph{Frames}: free-space observations that remove an obstacle observed many times. \emph{Error}, \emph{Real}: erroneous and real obstacle cells left at the end of an episode, among the cells that each map marks. \emph{Encl.}: frames in which the map encloses the agent. \emph{No frontier}: frames without a reachable frontier, among those in which the map from sensor depth has one.}
\label{tab:suppl_apex}
\end{table}

\paragraph{Enclosure.}
A stricter criterion counts the frames in which the map encloses the agent completely.
Only frames after which the agent actually moves at least $1$\,m away, within the next $64$ frames and $3$\,m of travel, are counted, so that the agent is known not to be enclosed in reality; this holds for $21{,}602$ frames.
The map encloses the agent in such a frame when, with obstacles enlarged by the agent radius, it has no path from the agent to that distance.
With predicted depth, this happens in $47\%$ of these frames for VLFM, $36\%$ for SG-Nav, $29\%$ for ApexNav, $26\%$ for SemExp, $18\%$ for the rebuilt map, $8.8\%$ for L3MVN and never for revisable occupancy, and in at most $2.1\%$ with sensor depth.
The map of L3MVN rarely encloses the agent because it clears what the current frame observes, but the erroneous obstacles that it keeps still leave the agent without a reachable frontier in most frames.

\paragraph{Erroneous and real cells.}
\Cref{fig:maprules}b and c follow obstacle cells over time.
An occupied cell is erroneous when the sensor depth at the same pose shows it to be free over the height of the agent, and real when the sensor depth places an obstacle point in it.
Each cell is followed from the step at which it is first labeled, and a cell that is cleared and later marked again counts as occupied.
At the end of an episode, the maps of VLFM, SemExp and SG-Nav keep all erroneous and real cells, the map of ApexNav $87\%$ and $100\%$, the map of L3MVN $56\%$ and $97\%$, the rebuilt map $24\%$ and $45\%$, and the revisable map $24\%$ and $89\%$.

\subsection{Exploration Statistics}
\label{sec:suppl_exploration}
\Cref{tab:suppl_exploration} shows how the systems of \cref{tab:generalization,tab:ablation} fail: whether the agent ever sees the goal, and whether the episode ends because no exploration goal is left.
Both quantities are computed from the ground truth after the episode; the goal counts as seen once at least $20$ of its pixels are visible in two consecutive frames.
With predicted depth, VLFM sees the goal in only $27\%$ of the episodes and runs out of exploration goals in $87\%$; with both of our components, these become $83\%$ and $0\%$.
Restricted to frontiers, \method runs out of exploration goals in $60\%$ of the episodes with predicted depth, against $23\%$ with sensor depth.

\input{tables/suppl_exploration}

\section{Depth, Mapping and Exploration: Details}
\label{sec:suppl_explore}
This section and the next give the details needed to reproduce \method; \cref{tab:suppl_params} lists the values of all parameters.

\input{tables/suppl_params}

\paragraph{Depth prediction.}
Each request contains the current image and up to five keyframes, chosen from the last $40$ steps for viewpoint diversity, recency and goal detections.
The model predicts metric scale with a dedicated branch, to which we pass the known intrinsics.
The camera poses enter the model only up to scale, and we do not rescale the prediction to the known poses afterwards, because this rescaling is undefined while the agent only turns in place, as it does at the start of every episode.
For the residual $r_t$ of \cref{eq:sigma}, the previous depth map is projected into the current view, keeping the nearest surface where several fall on the same pixel.
A current surface that is nearer than the projected one by more than $0.15\,\mathrm{m}+0.10\hat D_t$ may hide it and is not counted as a conflict; $r_t$ is zero where no comparison is possible.
$M_t$ is shrunk by one pixel at its border before object surfaces are built.
\input{fig/clearing_suppl}

\paragraph{Occupancy update.}
The grid is kept separately for each height layer, so that the floors of a multi-story scene do not overwrite each other.
Depth is subsampled by two in both image directions before integration, and a cell obtains an obstacle observation when at least two obstacle points fall into it.
Relative to the height of the agent's base, points between $-0.22$ and $0.15$\,m are ground, and points above the climbable step and below the agent's height are obstacle points.
Columns are grouped in pairs.
A group with at least three obstacle points casts a free ray to the $8\%$ quantile of their depths minus $0.08$\,m; otherwise, a group with at least three ground points casts a ray to the $85\%$ quantile of the ground depths.
Rays are three cells wide, and a cell that receives an obstacle observation in a frame is not cleared by the rays of the same frame.
Swept-footprint cells are reset to $e_m=-E$ after every update, and occupied cells are inflated by the agent radius of $0.18$\,m to obtain traversable space.
A forward action counts as failed when the agent moves less than $0.10$\,m; it blocks a disc of radius $0.10$\,m around the nominal landing position.
\Cref{fig:clearing_suppl} shows a second example of free-space clearing.

\paragraph{Exploration score.}
Candidates are the known-free cells with a route length between $0.5$ and $8$\,m, one per $0.4$\,m tile.
The gain and cost of \cref{eq:candidates} are
\begin{equation}
 \begin{gathered}
 G(v)=\lambda_u\,U(v)+\lambda_s\,V_g(v)+\lambda_r\,R(v),\\
 C(v)=P(v)-B(v),
 \end{gathered}
 \label{eq:score}
\end{equation}
where $U$ measures the distance-weighted visibility of unknown space, $V_g$ is the goal-conditioned image--text similarity stored in a value map as in VLFM~\cite{yokoyama2023vlfm}, $R$ is the region priority defined below, $B$ favors obstacle clearance and moderate displacement while penalizing long routes, and $P$ penalizes repeated visits and recently completed or abandoned goals.
The weights are $\lambda_u=1.8$, $\lambda_s=0.4$ and $\lambda_r=0.9$; the weights inside $B$ and $P$ are part of their definitions.
With $\rho_v$ the route length to $v$ in meters,
\begin{align*}
 U(v) &= \frac{1}{24}\sum_{k=1}^{24}\frac{\mathbf{1}[r_k\le 3\,\mathrm{m}]}{1+0.2\,r_k},\\
 B(v) &= 0.2\min(\mathrm{cl}(v),1)+0.25\min(\rho_v,2)-0.08\,\rho_v,\\
 P(v) &= 0.4\log(1+n_v)+0.4\log(1+h_v)+1.2\,f_v,
\end{align*}
where $r_k$ is the range at which ray $k$ from $v$ first reaches unknown space, and $r_k=\infty$ when the ray meets an occupied cell first.
$V_g$ and $R$ are min--max normalized over the candidate set, and $\mathrm{cl}(v)$ is the distance from $v$ to the nearest occupied cell in meters.
$n_v$ counts how often the agent entered the $0.75$\,m tile of $v$, $h_v$ counts the agent positions (recorded every $0.20$\,m of travel) within $0.65$\,m of $v$, and $f_v$ decays linearly from one to zero over $120$ steps after a goal in the tile of $v$ was last completed or abandoned.

\paragraph{Execution.}
The planner runs a beam search over $12$ discrete headings and $0.25$\,m forward moves on a distance field to the goal, penalizing turns, revisits and reversals of recent transitions.
A planned sequence is executed for at most $12$ steps while it makes progress.

\paragraph{Regions.}
Free space is divided into regions that do not cross walls and are separated at doorways, by growing them from the places whose distance to the nearest obstacle exceeds $0.6$\,m.
The priority of a region is its uncovered fraction, multiplied by one minus a suppression strength in $[0,1]$.
Suppression rises only with observed coverage, never with a missing detection; it decays at every step and is reset when new free space is observed or the goal is detected, so that no region is retired for good.

\section{Target Verification: Details}
\label{sec:suppl_target}

\subsection{Candidate Tracking}
\label{sec:holding}

\paragraph{Reliable surfaces.}
\label{sec:geometry}
Object surfaces are built only from predictions that are likely to be correct.
Projecting the previous depth map into the current view gives a residual $r_t$ wherever both views observe the same surface.
A pixel is reliable if its confidence is not among the lowest of the frame and its prediction is not contradicted by the previous one, and its uncertainty grows with depth and with the residual:
\begin{align}
 M_t &= V_t\wedge[C_t\geq q_{\kappa}(C_t)]\wedge\neg X_t,
 \label{eq:trust}\\
 \sigma_t &= \max(\alpha+\beta\hat D_t,\;r_t).
 \label{eq:sigma}
\end{align}
Here $V_t$ marks valid predictions, $q_{\kappa}$ is the $\kappa$-quantile over valid pixels, and $X_t$ marks pixels whose residual exceeds a depth-dependent tolerance; a surface that newly occludes the previous one is not counted as a contradiction.
The occupancy map does not use these quantities, for the reasons given in \cref{sec:map}.

\paragraph{Candidates.}
The surface estimate $P_c$ of a candidate $c=(b_c,P_c,\mathcal H_c)$ consists of the 3D points of its most recent reliable observation.
A goal mask is first split into depth-connected components using the reliable pixels of \cref{eq:trust}; neighboring pixels are linked when their depths differ by at most $0.10$\,m plus $2.5\%$ of the smaller depth.
Each resulting surface $S_t$ is matched to an existing candidate or starts a new one, and $S_{c,t}\subseteq S_t$ denotes the part of $S_t$ that is matched to candidate $c$.

\paragraph{Matching.}
We first test spatial overlap under a fixed $0.15$\,m radius.
If this fails, a recent candidate can still be matched when two conditions hold: at least half of the valid projections of its latest observation fall inside the new component mask, and at least half of the new surface points lie within an uncertainty-dependent radius of the current surface estimate,
\begin{equation}
 r_{c,t}=\operatorname{clip}\!\left(2\operatorname{median}_{u\in\Omega_{c,t}}\sigma_t(u),\;0.15,\;0.50\right),
 \label{eq:assoc}
\end{equation}
where $\Omega_{c,t}$ consists of the projected pixels of the latest observation that fall inside the new mask.

\paragraph{Score and position update.}
In \cref{eq:score_update}, $b_{\max}=4$ and $w=0.7$; the detector scores of the strongest detections of the goal category and of visually similar categories are mapped to bounded terms $\ell_c^+,\ell_c^-\in[0,2]$, and $a_c\in[-0.2,0.2]$ compares the similarity of the image patches of the surface to the goal text with their similarity to the texts of the similar categories.
The term $m_c$ penalizes a missing detection only when enough of the candidate is depth-consistent, uncropped and clearly in view.
A candidate is pursued once its score has been confirmed and it has passed simple geometric checks, which reject surfaces on the background plane and shapes that are implausible for the category; it is discarded when the score becomes negative.
The surface estimate $P_c$ is replaced only when a sufficiently large surface has been matched.
Earlier surfaces are thus not fused into the current estimate, but they remain available for the multi-view check.
If the pursued object is later split into a new candidate that overlaps the old one in the image, pursuit is transferred to the new candidate.

\subsection{Approaching and Stopping}
\label{sec:arrival}

\paragraph{Distance to the object surface.}
Instead of navigating to an object centroid or to its nearest isolated point, we measure the planar distance to a locally dense part of the surface.
For a point set $Q$, let $\Phi(Q)$ remove the points whose $12$-th neighbor in 3D is farther than $0.18$\,m, and define
\begin{equation}
 d(x;Q)=\operatorname{kth}_{p\in\Phi(Q)}\|x-p_{xy}\|_2,\qquad k=12,
 \label{eq:distance}
\end{equation}
with $d=+\infty$ when fewer than $12$ points remain.
Planning uses $d(x;P_c)$.
The stopping test uses $d_t=d(x_t;Q_{c,t})$, where $Q_{c,t}$ contains the points of the latest observation that remain depth-consistent after reprojection into the current frame, and $\bar\sigma_c$ denotes the $80\%$ quantile of $\sigma_t$ over these points.

\paragraph{Approach endpoint.}
We search for a reachable known-free endpoint satisfying
\begin{equation}
 d(x;P_c)<R-\mu_c-\epsilon,
 \label{eq:approach}
\end{equation}
with $R=0.9$\,m, $\epsilon=0.12$\,m and $\mu_c=\min(\bar\sigma_c,0.30\,\mathrm{m})$, or $\mu_c=0.15$\,m when too few points are available to compute $\bar\sigma_c$.
A more uncertain estimate therefore yields an endpoint closer to the object.
When no such endpoint is mapped yet, the planner selects a reachable intermediate endpoint and replans as observations arrive.

\paragraph{Keeping the object in view.}
For a currently detected object within $0.9$--$2$\,m, we select actions whose next poses keep the object visible: at least $12$ distinct pixels of the observed surface must remain inside the image with a border margin.
A forward move must also reduce the surface distance by at least $5$\,cm along a clear segment, and a turn must enable such a move.
Otherwise the agent returns to path following.

\paragraph{Stopping conditions.}
\label{sec:stop}
The stopping checks of \cref{sec:target} are the following.
\begin{itemize}[nosep,leftmargin=*]
  \item \emph{Identity}: the candidate is confirmed and detected in the current frame, with a score for the goal category that is no lower than for any similar category.
  \item \emph{Facing}: the observed surface lies within $16^\circ$ of the current heading.
  \item \emph{Geometry}: enough surface pixels, a fraction of conflicting pixels below $0.25$ in the detection mask, and $\bar\sigma_c<0.30$\,m.
  \item \emph{Multi-view}: an earlier observation from a camera position at least $0.18$\,m away; among its valid projected points, at least $12$ points and $35\%$ fall inside the current component mask and lie within $0.35$\,m of the current surface.
  \item \emph{Distance}: $d_t<\tau$ with $\tau=1.05$\,m.
\end{itemize}
When the geometry or multi-view condition is not met, the agent moves to a reachable position $0.25$--$0.65$\,m away, sideways, forward or backward depending on which condition failed, and turns back toward the object.
These moves, and attempts to detect the object again after it has been lost, are limited in steps; when a limit is reached, the agent returns to exploration.

\subsection{Analysis of Depth Errors and Failed Stops}
\label{sec:suppl_error}
\input{fig/suppl_deptherror}
Target verification makes two choices that follow from how the errors of predicted depth behave: the position of a candidate is replaced by its latest observation instead of being averaged over views, and the agent stops only on a close, current observation, which an observation from another view must confirm.
We check these choices against ground-truth depth and semantics rendered at the recorded poses of all HM3D v2 episodes ($6{,}460$ frames and $14{,}426$ observations of a candidate during the approach).

\paragraph{Depth error across views and distances.}
Most of the error of a frame is a common scale error: a single factor that scales the whole prediction accounts for $72\%$ of it in the median frame.
Nearby views of the same target share this error.
As \cref{fig:suppl_deptherror}b shows, the errors of two views whose camera positions are $0.18$--$0.5$\,m apart have a correlation of $0.87$, so averaging them removes only $3\%$ of the error, whereas it would remove $29\%$ if the errors were independent.
The error on the target surface is, in contrast, about $4\%$ of the distance at all distances, so that, as \cref{fig:suppl_deptherror}a shows, it shrinks from $18$\,cm at $3.5$--$4$\,m to $3.8$\,cm at $0.5$--$1$\,m as the agent approaches.
The reliability mask of \cref{eq:trust} removes most of the remaining bad pixels: in the median frame, $99.96\%$ of the pixels that it keeps are within a factor of $1.25$ of the true depth, against $76.3\%$ of the pixels that it rejects as conflicting.

\paragraph{Causes of failed stops.}
At every step at which the agent tests the stopping conditions within $\tau$ of a candidate, we check which object the observed surface of the candidate belongs to.
If it belongs to a goal instance, stopping there would fail in only $0.3\%$ of these steps; if it belongs to another object, in $78\%$.
Steps on a wrong object thus account for $97\%$ of the steps at which stopping would fail, and errors in the predicted distance for almost none.
The multi-view condition addresses these errors: near the goal, $11.5\%$ of the steps that pass the geometry and multi-view conditions are on a wrong object, against $32.7\%$ of the steps that fail them.
Of $856$ candidates observed near the goal, one changes from a surface on the goal to a surface on another object when its position is replaced by the latest observation.

\section{Additional Experiments}
\label{sec:suppl_more}

\subsection{Sensitivity to Odometry Drift}
\label{sec:suppl_drift}
All benchmark experiments use the poses provided by the simulator.
To examine how \method behaves when the poses drift, as wheel odometry does on a robot, we replace them on the $1{,}000$ HM3D v2 episodes by poses that accumulate noisy motion.
After every action, the pose passed to the system is updated with the true motion of the agent, in which the traveled distance is multiplied by $1+\sigma_d z_d$ and the rotation is offset by $\sigma_\theta z_\theta$, with $z_d, z_\theta\sim\mathcal{N}(0,1)$ drawn independently for every action.
The error accumulates over the episode and is never corrected.
The initial pose is exact, and SR and SPL are computed from the true positions.
\Cref{tab:suppl_drift} reports three noise levels together with the pose error they produce.
At the highest level, the pose is off by $0.96$\,m and $11.4^\circ$ on average at the end of an episode, and \method retains $95\%$ of the success rate obtained with exact poses.
The SR does not decrease monotonically with the noise level, and the SPL changes by at most $1.4$ points.

\begin{table}[h]
\centering
\small
\setlength{\tabcolsep}{4.5pt}
\begin{tabular}{@{}lcccccc@{}}
\toprule
& \multicolumn{2}{c}{Noise per action} & \multicolumn{2}{c}{Final pose error} & & \\
\cmidrule(lr){2-3}\cmidrule(lr){4-5}
Level & $\sigma_d$ & $\sigma_\theta$ & Position & Heading & SR & SPL \\
\midrule
None   & --    & --          & --       & --           & \res{v2}{1.0}{SR} & \res{v2}{1.0}{SPL} \\
Low    & 1\%   & $0.1^\circ$ & 0.10\,m  & $1.2^\circ$  & 67.3 & 30.1 \\
Medium & 2\%   & $0.3^\circ$ & 0.29\,m  & $3.4^\circ$  & 69.9 & 30.8 \\
High   & 5\%   & $1.0^\circ$ & 0.96\,m  & $11.4^\circ$ & 66.9 & 29.2 \\
\bottomrule
\end{tabular}
\caption{\textbf{Navigation with drifting poses} on HM3D v2. The final pose error is the error at the end of an episode, averaged over the episodes.}
\label{tab:suppl_drift}
\end{table}

\subsection{Runtime}
\label{sec:suppl_runtime}
During evaluation, several episodes run in parallel and share the models, which run as services on two NVIDIA A100 GPUs.
One step then takes on average $0.50$\,s for depth prediction, $0.45$\,s for perception, $0.13$\,s for updating the occupancy and value maps, and $0.41$\,s for planning.
These times include waiting for the shared services.
On the robot, where the models serve a single episode, one step takes about $0.7$\,s (\cref{sec:suppl_real}).

\subsection{Transfer to Existing Systems: Details}
\label{sec:suppl_transfer}
\input{fig/trajectories}
\paragraph{VLFM.}
All experiments with VLFM use its official implementation and weights, and the predicted depth comes from the same depth model and keyframe selection as in \method.
A simpler alternative, discussed in \cref{sec:map}, is to remove unreliable pixels before they reach VLFM: when the pixels rejected by $M_t$ (\cref{eq:trust}), on average $18\%$ of a frame, are treated as missing returns, the SR under predicted depth rises to $42.2\%$ and the SPL to $20.9\%$, below the $61.7\%$ and $30.2\%$ obtained with our two components.
Our two changes are applied on top of a relaxed termination: we disable VLFM's filter on small unexplored regions, and when no frontier is found, the agent first completes a full turn in place in $12$ steps and stops only if there is still none.
This alone raises the SR under predicted depth to $17.7\%$; adding revisable occupancy raises it to $50.7\%$, and adding exploration from reachable viewpoints to $61.7\%$.
For \emph{+ revisable occupancy}, we change only how VLFM's obstacle map integrates its observations.
Following \cref{sec:map}, an obstacle observation adds $\delta$ to the evidence of a cell and a free ray subtracts it, the evidence is bounded by $\pm E$ with the values of \cref{tab:suppl_params}, and the cells swept by the agent are reset to free.
VLFM's extraction of obstacle points, its inflation of obstacles and its frontier detection are unchanged.
For \emph{+ reachable viewpoints}, we intervene at a single point, namely where VLFM would stop because no frontier is left after the turn in place.
There, the exploration of \cref{sec:explore} selects a reachable viewpoint with the rules and weights of our system, using VLFM's own value map as the semantic term.
As soon as a frontier reappears, VLFM's frontier exploration takes over again, and the episode ends when no viewpoint has been available for the number of turns given in \cref{tab:suppl_params}.
VLFM's object detection, approach and stopping are the same in all variants, and the experiments with VGGT use the same changes.
\Cref{fig:trajectories} compares the trajectories of VLFM, VLFM with both components and \method on the same episodes.

\paragraph{L3MVN.}
We run the released zero-shot version of L3MVN~\cite{yu2023l3mvn}, with its semantic map, language-model scoring of frontiers and planner, on the same $1{,}000$ episodes and under the same actions, step limit and success criterion as all other experiments.
Predicted depth replaces the sensor in the map and in the RGB-D semantic segmentation of L3MVN, so its drop reflects the whole system.
L3MVN stops when its planner finds no path for $27$ consecutive steps, which we count as early termination.
L3MVN already clears cells that the current frame observes as free after its maximum fusion; our first change replaces this rule by revisable occupancy and leaves its semantic channels unchanged.
Our second change takes over only where L3MVN would stop for lack of a path and lets the agent continue toward a reachable viewpoint, chosen with L3MVN's own frontier scores and planner, until a goal of L3MVN becomes reachable again.
The experiments with VGGT use the same changes, and VGGT replaces the depth sensor in all components of L3MVN that consume depth.

\subsection{Depth-Model Substitution: Details}
\label{sec:suppl_vggt}
VGGT~\cite{wang2025vggt} predicts depth and camera poses up to an unknown scale, separately for each set of input images.
We recover the metric scale from the known camera height: a horizontal ground plane is fitted to the points of all input images, using the known camera orientations, and the scale is the ratio between the known camera height and the fitted distance from the camera to the ground.
When the images show no reliable ground, the scale is taken, in this order, from an image shared with an earlier set, from the ratio between known and predicted camera baselines of at least $2$\,m, or from a weak height prior.
No ground-truth depth is used.
\Cref{tab:suppl_depth} compares the two deployed depth sources on the same $316$ frames of $20$ episodes, against sensor depth rendered at the same poses.
These frames are sampled along trajectories recorded with sensor depth and are not the frames of \cref{sec:suppl_replay}.
With an ideal scale per frame, computed from the ground truth, the two models are comparably accurate, so the larger deployed error of VGGT comes from scale recovery.

\begin{table}[h]
\centering
\small
\setlength{\tabcolsep}{4.5pt}
\begin{tabular}{@{}lcccccc@{}}
\toprule
& \multicolumn{3}{c}{Deployed} & \multicolumn{3}{c}{Ideal scale} \\
\cmidrule(lr){2-4}\cmidrule(lr){5-7}
Depth model & Mean & Median & P90 & Mean & Median & P90 \\
\midrule
DA3  & 11.0 & 7.3 & 21.4 & 4.0 & 2.5 & 9.0 \\
VGGT & 16.7 & 9.8 & 30.3 & 3.3 & 2.6 & 5.2 \\
\bottomrule
\end{tabular}
\caption{\textbf{Relative depth error} (\%) of the two depth sources as deployed, and with an ideal per-frame scale. The error of a frame is the mean absolute relative error over its pixels; we report the mean, the median and the 90th percentile over frames.}
\label{tab:suppl_depth}
\end{table}

\subsection{Components of Target Verification}
\label{sec:suppl_target_ablation}
\Cref{tab:suppl_target} decomposes the target-verification ablation of \cref{tab:ablation} into its three parts: matching detections to candidates by image overlap and replacing their position (\emph{position update}), keeping the object in view during the final approach together with the extra observation moves (\emph{view-keeping approach}), and the geometric-consistency and multi-view conditions before stopping (\emph{stopping checks}).
The position update accounts for most of the joint effect, whereas the other two parts have small individual effects, which is why \cref{tab:ablation} evaluates target verification as a whole.
The row without stopping checks is obtained by replaying the stop decision on the recorded episodes of the full system.

\begin{table}[h]
\centering
\small
\begin{tabular}{@{}lccc@{}}
\toprule
Variant & SR & SPL & $\Delta$SR \\
\midrule
\method (full) & \res{v2}{1.0}{SR} & \res{v2}{1.0}{SPL} & -- \\
\quad w/o target verification & 67.6 & 29.9 & $-2.5$ \\
\quad w/o position update & 68.2 & 30.6 & $-1.9$ \\
\quad w/o view-keeping approach & 69.3 & 30.4 & $-0.8$ \\
\quad w/o stopping checks & \abl{A2S}{SR} & \abl{A2S}{SPL} & $-0.3$ \\
\bottomrule
\end{tabular}
\caption{\textbf{Components of target verification} on HM3D v2. \emph{w/o target verification} removes all three parts; each of the last three rows removes one of them.}
\label{tab:suppl_target}
\end{table}

\section{Real-World Deployment: Details}
\label{sec:suppl_real}
\input{fig/realworld_full}

The AgiBot Genie G1 has a wheeled base and a humanoid upper body.
We use the RGB camera in its head at a resolution of $1280\times800$ with a horizontal field of view of about $89^\circ$; in the posture used for navigation it is $1.15$\,m above the floor and pitched $22^\circ$ downward.
The poses are those of the wheel odometry of the robot and are not corrected by any other sensor.
As in simulation, the actions are a forward step of $0.25$\,m and turns of $30^\circ$, each executed by a velocity controller on the odometry.
All models run on a workstation that is connected to the robot by a network cable, and one step of perception, depth prediction and planning takes about $0.7$\,s.
So that the cable does not wind up, the robot covers the headings of its initial turn in place by turning in both directions.
\Cref{tab:suppl_real} lists the twelve runs that we report, with chairs, trash cans and potted plants as targets; trash cans are not among the categories of the simulation benchmarks.
A run is successful when the robot stops within $1$\,m of the target, measured from the center of the robot.
\Cref{fig:realworld_full} shows runs 1 and 4.

\begin{table}[h]
\centering
\small
\setlength{\tabcolsep}{4pt}
\begin{tabular}{@{}clcccc@{}}
\toprule
Run & Target & Start & In view & Path & Stop \\
\midrule
1  & chair     & 6.5\,m & \xmark & 6.1\,m & 0.7\,m \\
2  & chair     & 4.3\,m & \xmark & 5.4\,m & 0.7\,m \\
3  & trash can & 1.6\,m & \cmark & 1.7\,m & 0.7\,m \\
4  & trash can & 1.6\,m & \xmark & 1.2\,m & 0.7\,m \\
5  & chair     & 3.7\,m & \cmark & 3.4\,m & 0.7\,m \\
6  & chair     & 3.6\,m & \xmark & 2.7\,m & 1.0\,m \\
7  & trash can & 4.0\,m & \xmark & 5.6\,m & 0.8\,m \\
8  & trash can & 5.3\,m & \cmark & --      & no stop \\
9  & plant     & 3.8\,m & \cmark & 3.0\,m & 0.8\,m \\
10 & plant     & 4.6\,m & \cmark & 5.8\,m & 0.6\,m \\
11 & plant     & 2.9\,m & \xmark & 2.5\,m & 0.9\,m \\
12 & plant     & 5.1\,m & \xmark & 8.2\,m & 0.9\,m \\
\bottomrule
\end{tabular}
\caption{\textbf{Real-robot runs.} Start: distance from the start position to the target. In view: target visible in the first image. Stop: distance from the center of the robot to the target when it stops; in run 8 the stopping conditions were never met.}
\label{tab:suppl_real}
\end{table}

%% file: fig/suppl_replay.tex
\begin{figure*}[t]
  \centering
  \includegraphics[width=\linewidth]{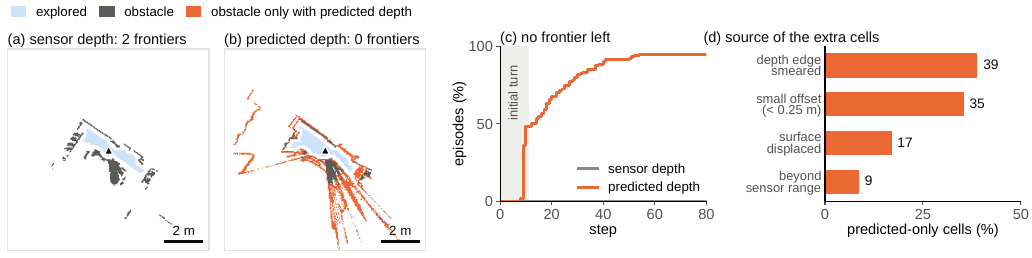}
  \caption{\textbf{The map of VLFM at identical poses with sensor and with predicted depth.}
    (a, b) The two maps of one episode after the initial turn.
    (c) Episodes in which the map has no frontier left, over the $120$ episodes.
    (d) Obstacle cells that only the map from predicted depth contains, by the sensor reading at the pixels that produce them.}
  \label{fig:suppl_replay}
\end{figure*}

%% file: fig/suppl_maprules.tex
\begin{figure*}[t]
  \centering
  \includegraphics[width=\linewidth]{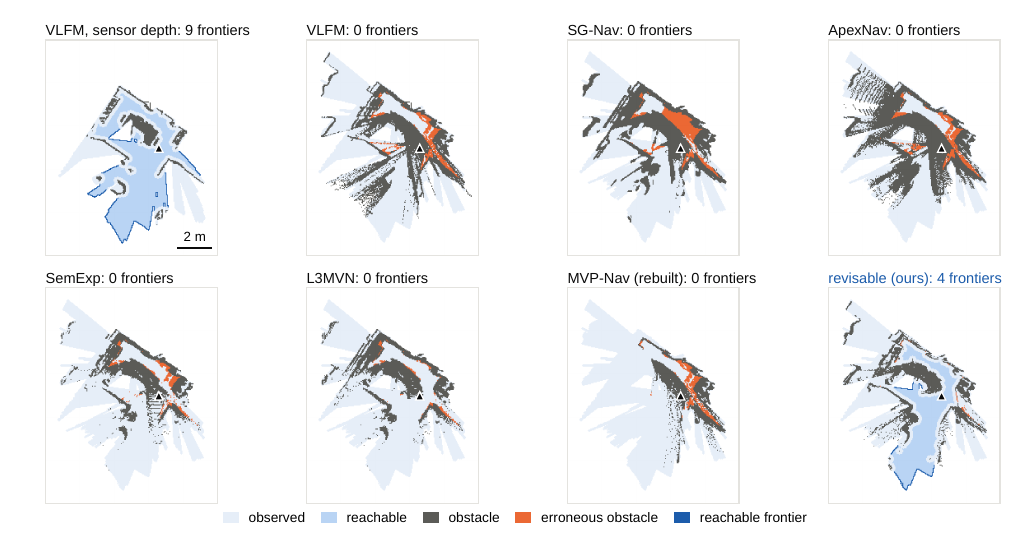}
  \caption{\textbf{The seven update rules at the same frame.}
    All maps except the first are built from the same predicted depth along the same trajectory; the first is the map of VLFM from sensor depth.
    Blue shows the region that the agent can reach and its reachable frontiers.}
  \label{fig:suppl_maprules}
\end{figure*}

%% file: tables/suppl_exploration.tex
\begin{table}[h]
\centering
\small
\setlength{\tabcolsep}{3.2pt}
\begin{tabular}{@{}llcccc@{}}
\toprule
System & Depth & SR & SPL & Seen & Exh. \\
\midrule
VLFM                          & sensor & 68.2 & 34.5 & 88.4 & 8.0 \\
VLFM                          & DA3    & 10.8 & 7.5  & 27.1 & 86.9 \\
\quad + relaxed termination   & DA3    & 17.7 & 11.0 & 36.9 & 74.2 \\
\quad + revisable occupancy   & DA3    & 50.7 & 25.8 & 70.0 & 24.8 \\
\quad + reachable viewpoints  & DA3    & 61.7 & 30.2 & 83.3 & 0.0 \\
VLFM                          & VGGT   & 8.3  & 6.1  & 25.4 & 88.6 \\
\quad + both components       & VGGT   & 59.8 & 31.5 & 83.9 & 0.1 \\
\midrule
\method                       & sensor & 70.7 & 33.7 & 86.8 & 0.1 \\
\quad frontier goals only     & sensor & 66.2 & 33.9 & 82.1 & 22.7 \\
\method                       & DA3    & \res{v2}{1.0}{SR} & \res{v2}{1.0}{SPL} & 84.7 & 0.3 \\
\quad w/o free-space clearing & DA3   & 13.9 & 6.4  & 54.8 & 1.9 \\
\quad frontier goals only     & DA3    & 32.6 & 18.1 & 49.3 & 60.1 \\
\method                       & VGGT   & 69.5 & 30.5 & 85.2 & 0.1 \\
\bottomrule
\end{tabular}
\caption{\textbf{Exploration statistics} on HM3D v2. \emph{Seen}: episodes in which the goal becomes clearly visible (\%). \emph{Exh.}: episodes that end because the agent has no exploration goal left (\%).}
\label{tab:suppl_exploration}
\end{table}

%% file: tables/suppl_params.tex
\begin{table}[h]
\centering
\footnotesize
\setlength{\tabcolsep}{4pt}
\begin{tabular}{@{}llc@{}}
\toprule
Parameter & Symbol & Value \\
\midrule
\multicolumn{3}{@{}l}{\emph{Depth and reliability}} \\
Keyframes per query            & --                       & up to 5 \\
Keyframe window                & --                       & 40 steps \\
Confidence quantile            & $\kappa$                 & 0.15 \\
Uncertainty offset and slope   & $\alpha$, $\beta$        & 0.03\,m, 0.05 \\
Conflict tolerance             & --                       & $0.10\,\mathrm{m}+0.08\hat D_t$ \\
\midrule
\multicolumn{3}{@{}l}{\emph{Revisable occupancy}} \\
Cell size                      & --                       & 5\,cm \\
Height layer                   & --                       & 0.5\,m \\
Evidence bound and step        & $E$, $\delta$            & 3, 1.5 \\
Obstacle points per cell       & --                       & $\geq 2$ \\
Ray margin before obstacles    & --                       & 0.08\,m \\
Agent radius                   & --                       & 0.18\,m \\
Blocked radius after a failed move & --                   & 0.10\,m \\
\midrule
\multicolumn{3}{@{}l}{\emph{Exploration}} \\
Route length of candidates     & --                       & 0.5--8\,m \\
Candidate spacing              & --                       & 0.4\,m \\
Visibility rays (number, range) & --                      & 24, 3\,m \\
Gain weights                   & $\lambda_u$, $\lambda_s$, $\lambda_r$ & 1.8, 0.4, 0.9 \\
Goal reached within            & --                       & 0.3\,m \\
No-progress limit              & --                       & 24 steps \\
Goal time limit                & --                       & 96 steps \\
Turns before termination       & --                       & 36 \\
\midrule
\multicolumn{3}{@{}l}{\emph{Target verification}} \\
3D matching radius             & --                       & 0.15\,m \\
Score bound                    & $b_{\max}$               & 4 \\
Detection weight               & $w$                      & 0.7 \\
Approach radius                & $R$                      & 0.9\,m \\
Stopping distance              & $\tau$                   & 1.05\,m \\
Facing tolerance               & --                       & $16^\circ$ \\
Multi-view baseline            & --                       & 0.18\,m \\
\midrule
\multicolumn{3}{@{}l}{\emph{Models}} \\
Input width (depth, perception) & --                      & 504, 672\,px \\
Detection threshold            & --                       & 0.4 \\
\bottomrule
\end{tabular}
\caption{\textbf{Parameter values} of \method.}
\label{tab:suppl_params}
\end{table}

%% file: fig/clearing_suppl.tex
\begin{figure}[t]
  \centering
  \includegraphics[width=\linewidth]{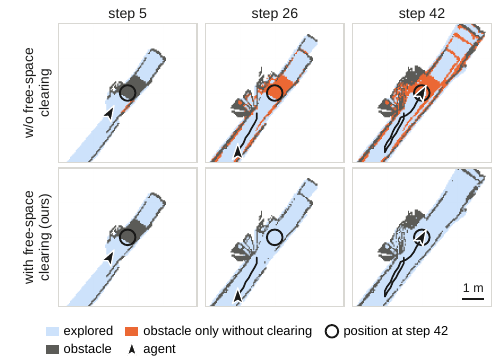}
  \caption{\textbf{A second example of free-space clearing}, in a corridor. The layout is the same as in \cref{fig:clearing}.}
  \label{fig:clearing_suppl}
\end{figure}

%% file: fig/suppl_deptherror.tex
\begin{figure}[t]
  \centering
  \includegraphics[width=\linewidth]{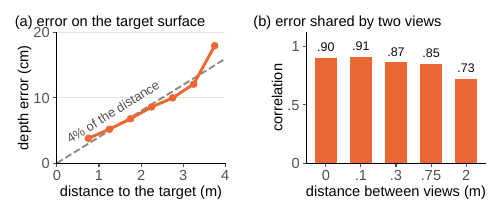}
  \caption{\textbf{Depth error on the target surface} while the agent approaches a candidate on HM3D v2.
    (a) Median error by the true distance to the target.
    (b) Correlation between the errors of two views of the same target, by the distance between the two camera positions; a correlation near one means that the second view repeats the error of the first.}
  \label{fig:suppl_deptherror}
\end{figure}

%% file: fig/trajectories.tex
\begin{figure*}[t]
  \centering
  \includegraphics[width=\linewidth]{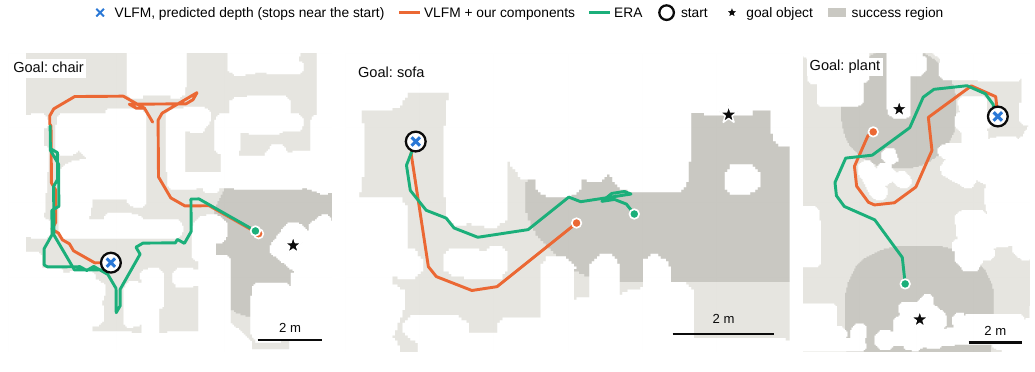}
  \caption{\textbf{Qualitative comparison} on HM3D v2 with predicted depth.
    VLFM stops within a few steps of the start because its map has no frontier left.
    With revisable occupancy and exploration from reachable viewpoints, it explores the scene and reaches the goal, as does \method.}
  \label{fig:trajectories}
\end{figure*}

%% file: fig/realworld_full.tex
\begin{figure*}[t]
  \centering
  \includegraphics[width=\linewidth]{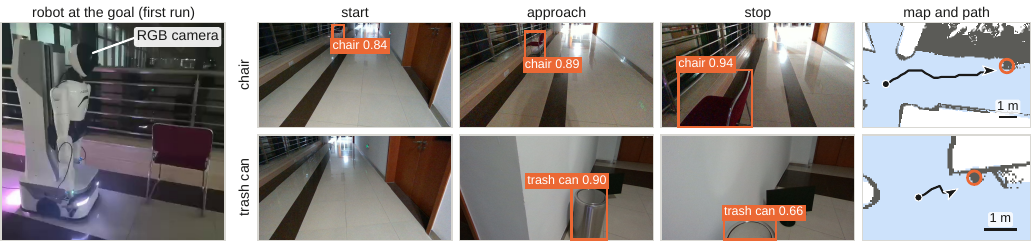}
  \caption{\textbf{Two runs on the robot.}
    Left: the robot at the goal of the first run; the RGB camera in its head is the only camera used.
    Right: the view of this camera at the start, during the approach and at the stop, where the boxes are the detections of the goal object, and the map built from predicted depth with the path of the robot (black) and the goal (orange circle).
    The robot stops $0.7$\,m from both the chair and the trash can; the trash can is not in view at the start.}
  \label{fig:realworld_full}
\end{figure*}